\RequirePackage{amsthm}
\documentclass[sn-mathphys-num]{sn-jnl}

\usepackage{geometry} 

\UseRawInputEncoding
\usepackage{graphicx}
\usepackage{subfigure}
\usepackage{multirow}%
\usepackage{amsmath,amssymb,amsfonts}%
\usepackage{amsthm}%
\usepackage{mathrsfs}%
\usepackage[title]{appendix}%
\usepackage{xcolor}%
\usepackage{textcomp}%
\usepackage{manyfoot}%
\usepackage{booktabs}%
\usepackage{algorithm}%
\usepackage{algorithmicx}%
\usepackage{algpseudocode}%
\usepackage{listings}%
\RequirePackage{geometry}
\usepackage{hyperref}  
\hypersetup{
	hidelinks,
	colorlinks=true,
	allcolors=black,
	pdfstartview=Fit,
	breaklinks=true
}
\usepackage[all]{hypcap}

\usepackage[]{indentfirst}
\usepackage{caption}
\usepackage{amsthm}%
\usepackage{lmodern}

\usepackage{textcomp}
\usepackage{mathrsfs}

\DeclareFontFamily{U}{rsfs}{\skewchar\font127 }
\DeclareFontShape{U}{rsfs}{m}{n}{%
	<-6.5> rsfs5
	<6.5-8> rsfs7
	<8-> rsfs10
}{}

\theoremstyle{thmstyleone}%
\theoremstyle{thmstyletwo}%

\theoremstyle{thmstylethree}%

\begin{document}

\title{MobileNetV2: A lightweight classification model for home-based sleep apnea screening}


\author[1]{\fnm{Hui} \sur{Pan}}\email{2310247001@email.szu.edu.cn}

\author[1]{\fnm{Yanxuan} \sur{Yu}}\email{yy3523@columbia.edu}

\author*[1,2]{\fnm{Jilun} \sur{Ye}}\email{yejilun@126.com}

\author*[1]{\fnm{Xu} \sur{Zhang}}\email{xuzhang@szu.edu.cn}


\affil[1]{\orgdiv{School of Medicine}, \orgname{Shenzhen University}, \orgaddress{\street{Xueyuan Avenue 1066}, \city{Shenzhen}, \postcode{518055}, \state{Guangdong Province}, \country{China}}}

\affil[2]{\orgname{Witleaf Medical Electronics Co.}, \orgaddress{\street{Guangming District}, \city{Shenzhen}, \postcode{518132}, \state{Guangdong Province}, \country{China}}}

%


\abstract{
	Obstructive Sleep Apnea (OSA) is a prevalent sleep disorder linked to severe complications such as hypertension, stroke, and epilepsy, posing significant health risks. Despite its high prevalence, early diagnosis of OSA remains challenging due to the high cost of screening and limited accessibility to specialized sleep centers, leading to a large number of undiagnosed cases. While neural networks have demonstrated high classification accuracy in OSA screening, their large size often limits their applicability to wearable devices, constraining practical use.
	
	This study proposes a novel lightweight neural network model leveraging features extracted from electrocardiogram (ECG) and respiratory signals for early OSA screening. ECG signals are used to generate feature spectrograms to predict sleep stages, while respiratory signals are employed to detect sleep-related breathing abnormalities. By integrating these predictions, the method calculates the apnea-hypopnea index (AHI) with enhanced accuracy, facilitating precise OSA diagnosis.
	
	The method was validated on three publicly available sleep apnea databases: the Apnea-ECG database, the UCDDB dataset, and the MIT-BIH Polysomnographic database. Results showed an overall OSA detection accuracy of 0.978, highlighting the model's robustness. Respiratory event classification achieved an accuracy of 0.969 and an area under the receiver operating characteristic curve (ROC-AUC) of 0.98. For sleep stage classification, in UCDDB dataset, the ROC-AUC exceeded 0.85 across all stages, with recall for Sleep reaching 0.906 and specificity for REM and Wake states at 0.956 and 0.937, respectively.
	
	This study underscores the potential of integrating lightweight neural networks with multi-signal analysis for accurate, portable, and cost-effective OSA screening, paving the way for broader adoption in home-based and wearable health monitoring systems.}




\maketitle

\section{Introduction}\label{sec1}
Obstructive sleep apnea(OSA) is a common sleep disorder characterized by repeated cessation of breathing during sleep\cite{noauthor_1999-ly}. The prevalence of OSA in the population ranges from 9\% to 38\% and increases with age, with the probability of moderate to severe OSA being as high as 49\% in the senior population\cite{Senaratna2017-sw}. OSA is emerging as a major health problem that cause complications like hypertension, stroke, diabetes and epilepsy\cite{Lloyd2024-oq,Chervin2000-ua}. Despite its high prevalence, OSA has reported to be undiagnosed in 85\% of potential patients\cite{Motamedi2009-bq}, due to the high costs of screening and the limited availability of sleep centers. Traditional PSG screening method is performed in sleep centers with sensors including electroencephalogram(EEG), electrocardiogram (ECG), nasal airflow, pulse oximeter(SpO2), respiratory effort chest belts, and others. Therefore, the development of a portable, easy-to-use, and cost-effective rapid screening tool for OSA should be a current priority.

Research has shown that convolutional neural networks (CNNs) provide an effective solution for interpreting and learning from physiological signal sequences, as deep learning algorithms can automatically extract complex features from raw data, enhancing both diagnostic accuracy and robustness\cite{Mostafa2019-cg}. Since AlexNet?s success in the 2012 ImageNet Challenge\cite{Krizhevsky2017-ib}, deep convolutional networks have trended towards increasing depth and structural complexity\cite{Simonyan2014-cj,He2015-aa}. While this can improve model accuracy, it often results in larger network sizes and slower computation. Due to hardware and computational limitations, complex deep learning models are challenging to deploy on mobile devices.

In response, this study proposes a lightweight deep convolutional approach for sleep apnea monitoring and validates the model's performance using three publicly available datasets.

\section{Materials and Methods}\label{sec2}
\subsection{Dataset}

The proposed method was validated using three publicly available datasets:

1. Apnea-ECG Database\cite{Penzel2000-fd}: Contains 70 ECG recordings with a sampling rate of 100 Hz. Selected records (a01-a04, b01, c01-c03) include both ECG and blood oxygen data.

2. UCDDB Dataset\cite{McNicholas2004-tm}: Comprises overnight PSG data from 25 patients with a sampling rate of 128 Hz, used for sleep stage differentiation.

3. MIT-BIH Polysomnographic Database\cite{Ichimaru1992-fj}: Includes 18 multi-channel recordings with a sampling rate of 128 Hz, broadening the model's predictive scope.

\subsection{Data preprocessing}

\begin{figure}[!htb]
	
	\centering
	\includegraphics[width=0.6\columnwidth]{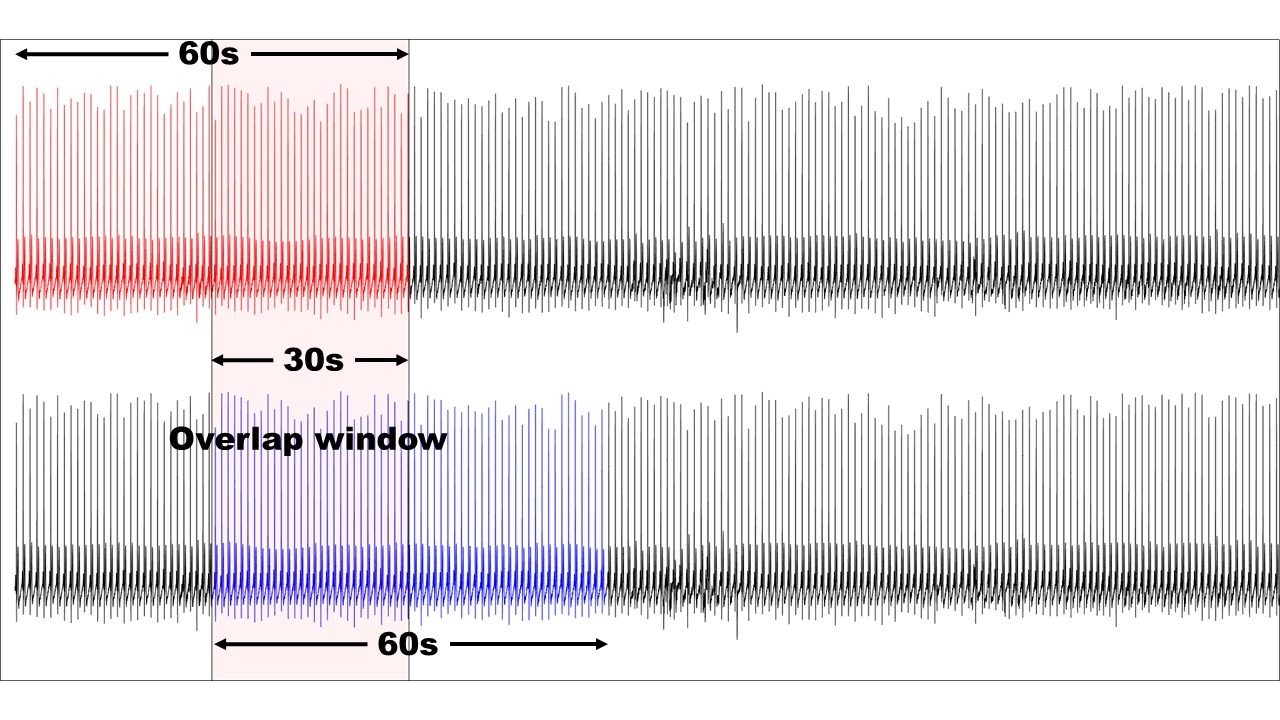}
	
	\caption{preprocess for ECG signal}\label{fig0_ecg_epoch}
\end{figure}

As reported in the literature\cite{De_Chazal2004-hm}, 1-minute data segments are more advantageous for screening respiratory events, while sleep stages are typically divided into 30-second segments. In this study, ECG data is uniformly divided into 1-minute segments, with a sliding window of 30 seconds, like Fig \ref{fig0_ecg_epoch}. This approach encompasses sleep states while preventing respiratory events from being split due to the segmentation. To more accurately calculate the AHI index, sleep periods in this study are classified into two states: Sleep (S) and Non-Sleep (NS). Segments labeled as N1, N2, N3, and N4 are considered as Sleep (S) states.

\begin{figure}[h]
	
	\centering
	\includegraphics[width=0.6\columnwidth]{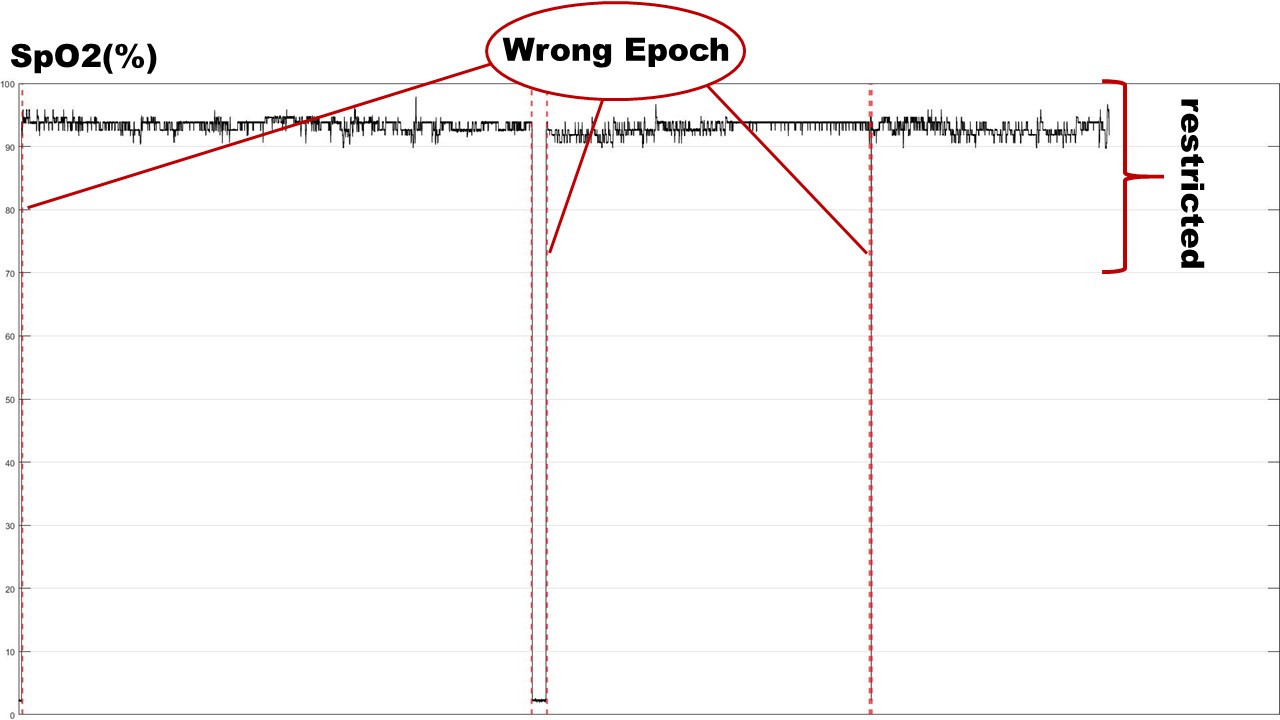}
	\caption{Overnight SpO2 data for subject 19 in the UCDDB database}
	\label{fig1_badsp}
\end{figure}

After re-segmenting the signals, we applied bandpass filter (0.3-45Hz) to the single-lead ECG to remove baseline offsets and power disturbances.Observing fluctuations and artifacts in the signals in two datasets, we applied the following two treatments to the ECG and SpO2, respectively:

For SpO2, it in healthy humans is chronically stable at 97\% or higher, while in critically ill patients undergoing surgery it is between 80\% and 100\%\cite{Hedenstierna2000-hr}. Therefore, we only retained the 60-second epochs with SpO2 in the 70\% to 100\% range. You can see those non-compliant SpO2 data in Fig \ref{fig1_badsp}.

\begin{figure}[h]
	
	\centering
	\subfigure[]
	{\includegraphics[width=0.4\columnwidth]{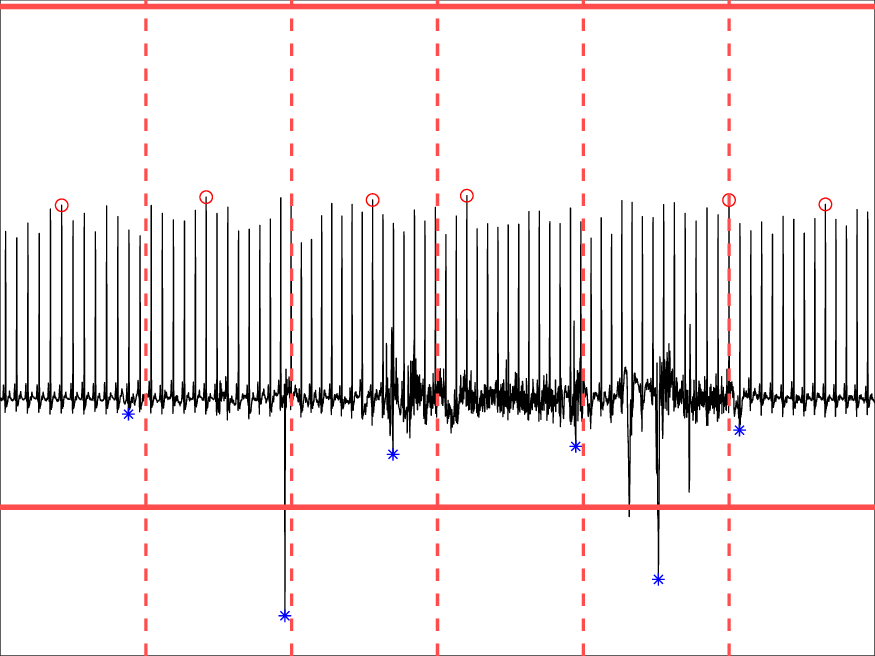}}
	\hfil
	\subfigure[]
	{\includegraphics[width=0.4\columnwidth]{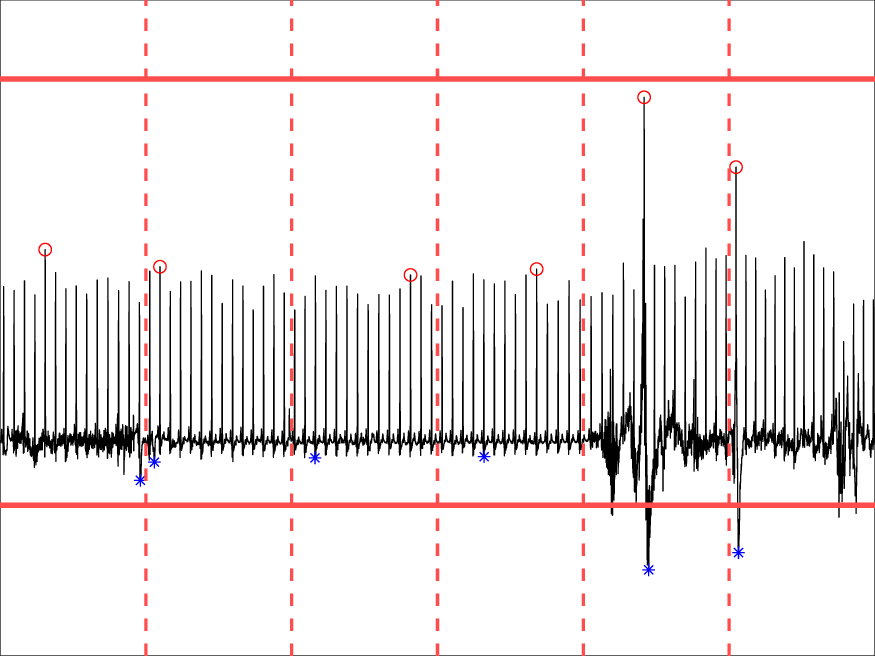}}
	\hfil \\
	
	\subfigure[]
	{\includegraphics[width=0.4\columnwidth]{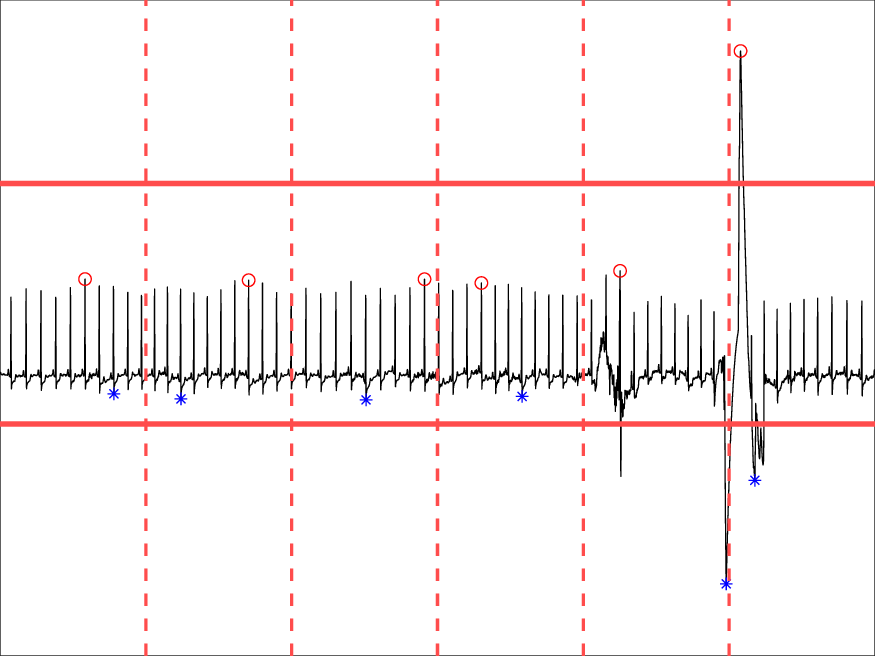}}
	\hfil
	\subfigure[]
	{\includegraphics[width=0.4\columnwidth]{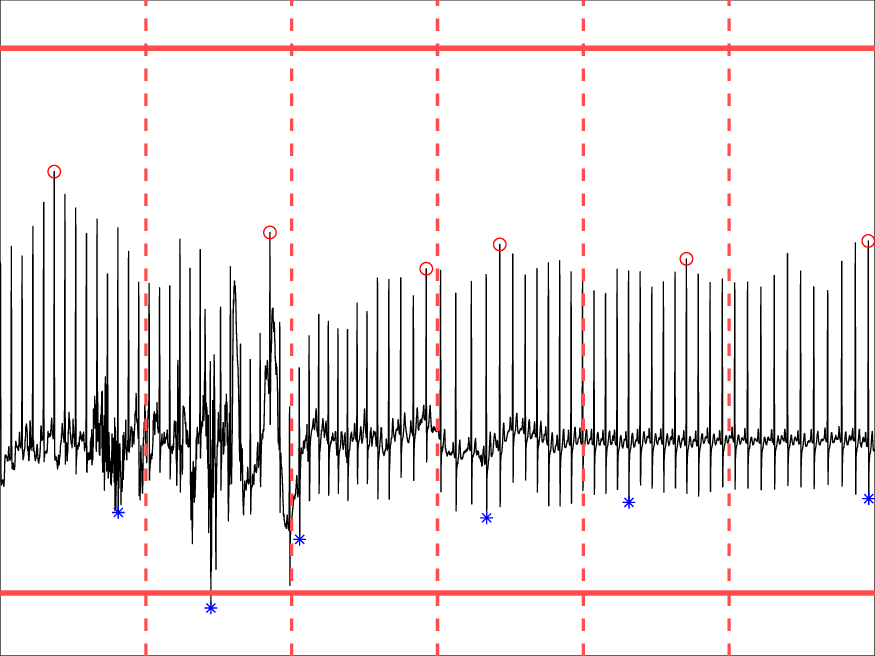}}
	
	\caption{Four examples of contaminated ECG segments, where: the red dot represents the maximum value of the 10-second segment and the blue asterisk represents the minimum value of the 10-second segment.}
	\label{fig_badcase}
\end{figure}

For ECG, we divided each 60-second segment into six 10-second subsegments and extracted the maximum and minimum values from each. The median of the six maximum values was set as threshold th1, and the median of the six minimum values as threshold th2. If any value within a subsegment exceeds 2*th1 or falls below 2*th2, the subsegment is considered contaminated. The contaminated 60-second segments are given in Fig \ref{fig_badcase}.

\begin{figure}[h]
	\centering
	\subfigure[ECG segment of 'A']
	{\includegraphics[width=0.4\columnwidth]{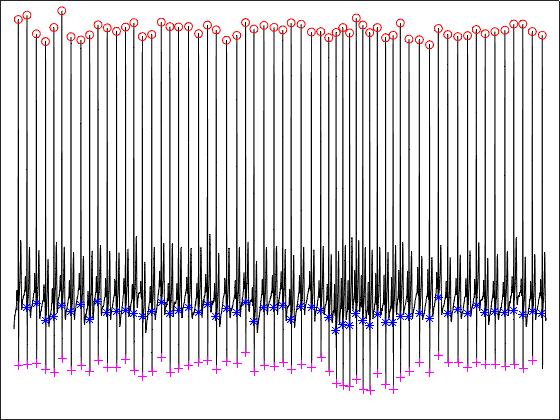}}
	\hfil
	\subfigure[ECG segment of 'N']
	{\includegraphics[width=0.4\columnwidth]{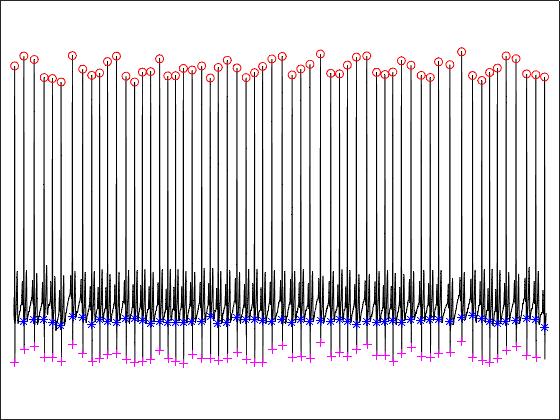}}
	
	\caption{Two examples of ECG segments with extracted QRS points.}
	\label{fig_ecgqrs}
\end{figure}

\begin{figure}[h]
	
	\centering
	\subfigure[eg1: Comparison plot of EDR and respiratory signals for 'A']
	{\includegraphics[width=0.4\columnwidth]{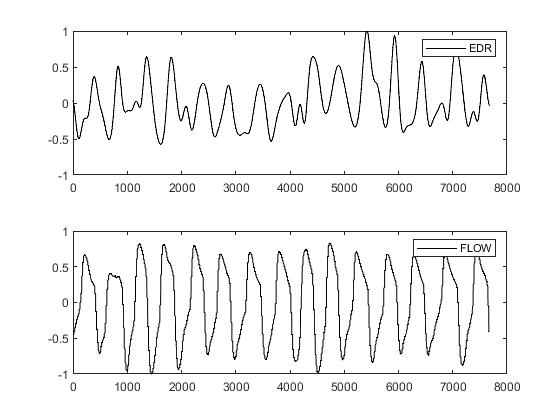}}
	\hfil
	\subfigure[eg2: Comparison plot of EDR and respiratory signals for 'A']
	{\includegraphics[width=0.4\columnwidth]{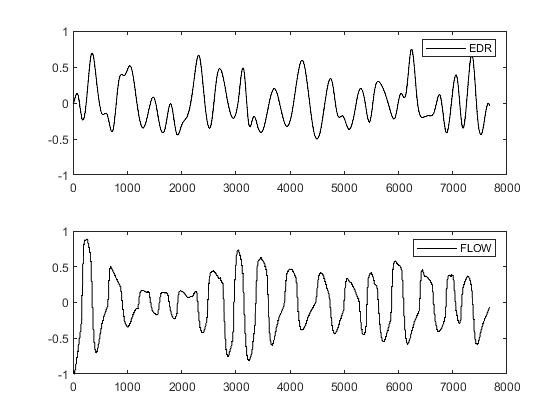}}
	\hfil \\
	
	\subfigure[eg1: Comparison plot of EDR and respiratory signals for 'N']
	{\includegraphics[width=0.4\columnwidth]{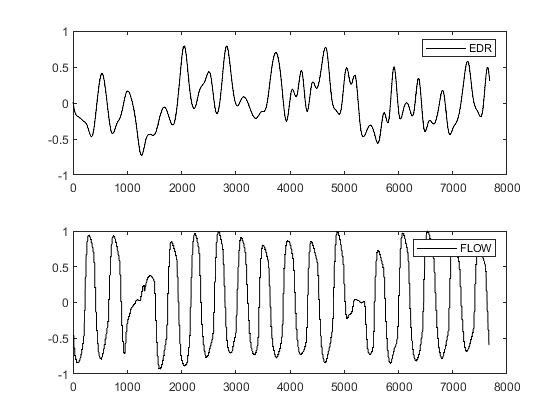}}
	\hfil
	\subfigure[eg2: Comparison plot of EDR and respiratory signals for 'N']
	{\includegraphics[width=0.4\columnwidth]{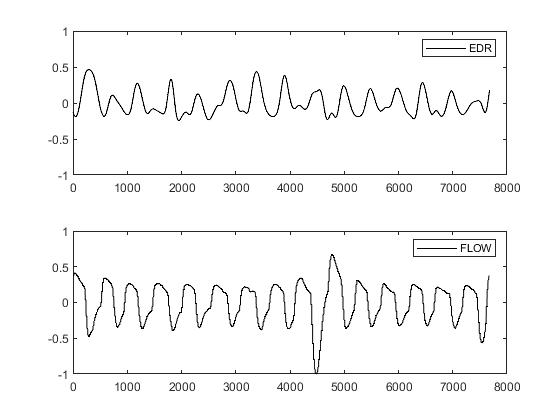}}
	
	\caption{Four images are provided to compare the synchronization between respiratory waveforms and EDR signals. (a) and (b) illustrate comparisons for abnormal respiratory events, while (c) and (d) show comparisons for normal respiratory events.}
	\label{fig_EDR}
\end{figure}

\begin{figure}[h]
	\centering
	\includegraphics[width=0.8\columnwidth]{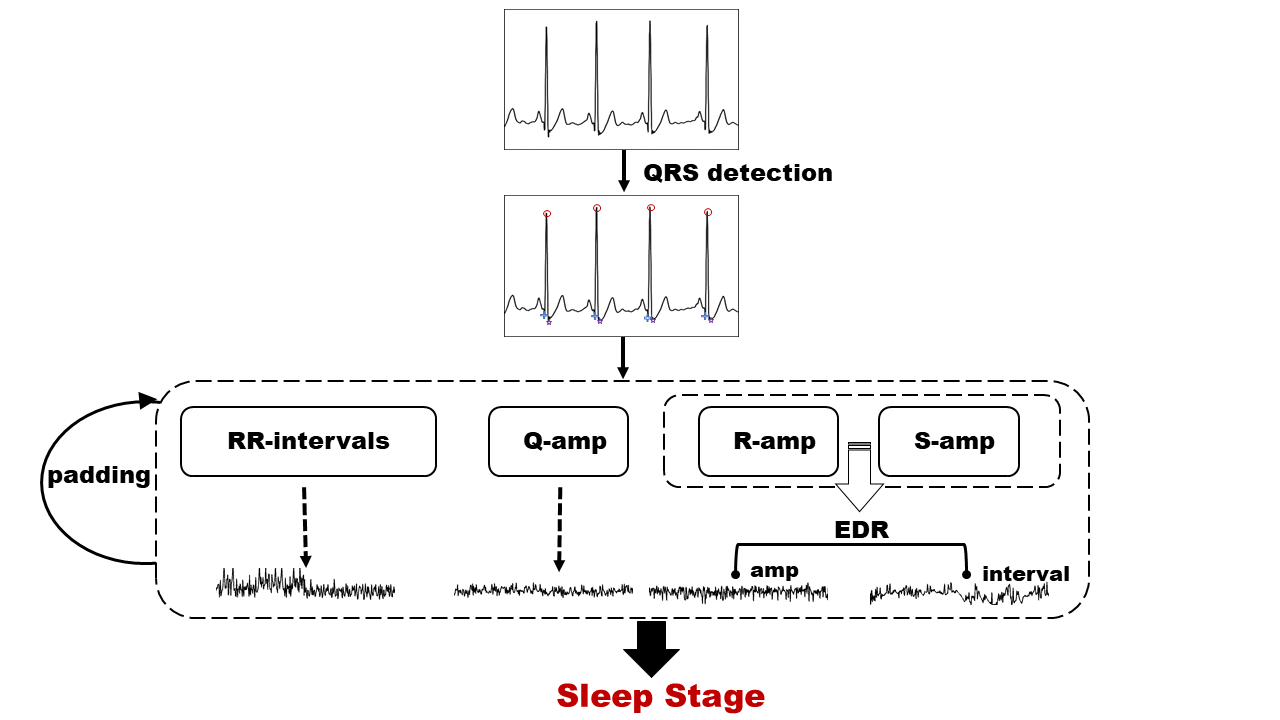}
	\caption{Flowchart of the algorithm for sleep staging using ECG signals}
	\label{fig_ecgalgo}
\end{figure}

Studies indicate that the respiratory-derived signal (EDR) extracted from ECG closely resembles the original respiratory signal\cite{Sarkar2015-dm,Varon2020-wr}. During sleep, events like restricted or paused breathing cause marked rhythmic changes in the RR interval of ECG signals\cite{Roebuck2014-yp}. Building on this, we extracted QRS feature points from clean ECG signals. Fig \ref{fig_ecgqrs} illustrates ECG signals under normal conditions and during respiratory events, while Fig \ref{fig_EDR} compares actual respiratory signals with simulated EDR signals. Given ECG?s ability to capture both respiratory and sleep information, we populated sequences for RR interval, Q-wave amplitude, R-wave amplitude, EDR amplitude, and EDR peak intervals to lengths of 200, 200, 200, 100, and 100 points, respectively. As shown in Fig \ref{fig_ecgalgo}, these populated feature sequences were input into a neural network model for sleep stage classification.

\subsection{Model architecture}
\begin{figure}[h]
	\centering
	\includegraphics[width=0.8\columnwidth]{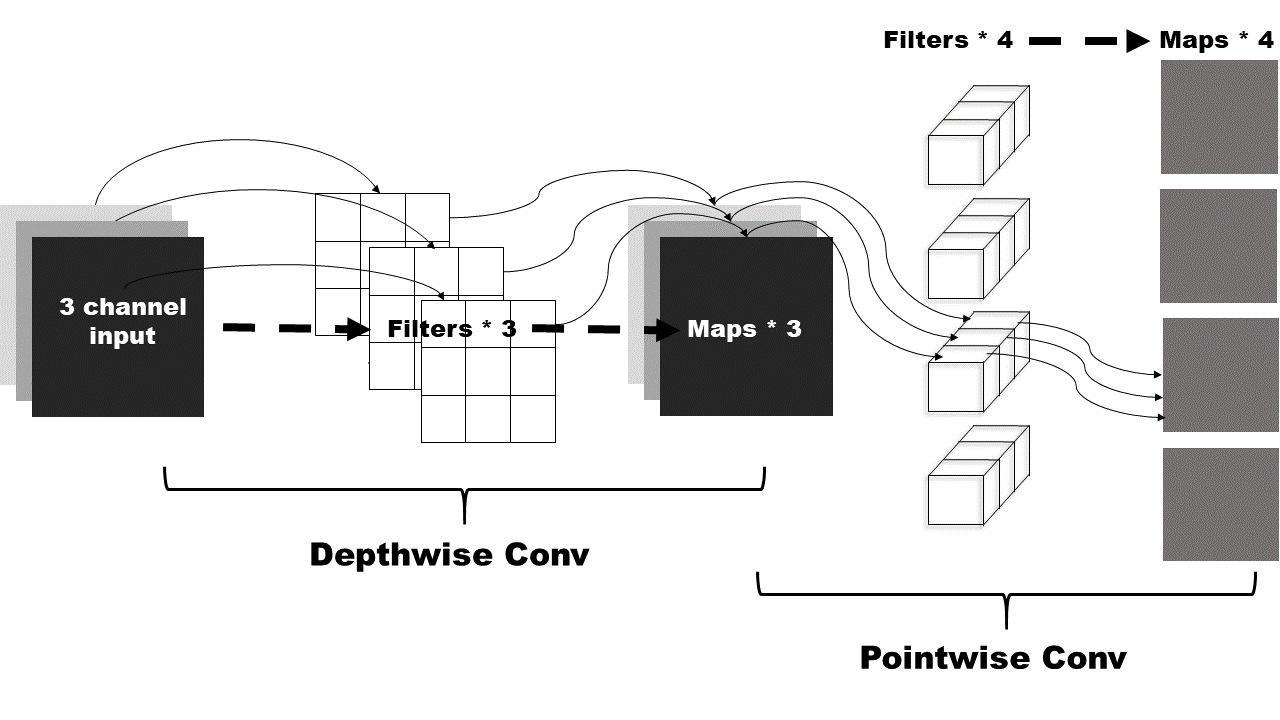}
	\caption{Example plot of deep convolution and point-by-point convolution for feature extraction}
	\label{fig_DP}
\end{figure}

The MobileNet model is based on depth-separable convolution, which is a form of decomposed convolution. This form decomposes the standard convolution into a depth convolution and a 1x1 convolution called point-by-point convolution. Fig \ref{fig_DP} illustrates the difference in feature extraction between depthwise and pointwise convolutions. This design reduces parameters and computational load, making it well-suited for mobile applications.

\begin{figure}[h]
	\centering
	\includegraphics[width=0.9\columnwidth]{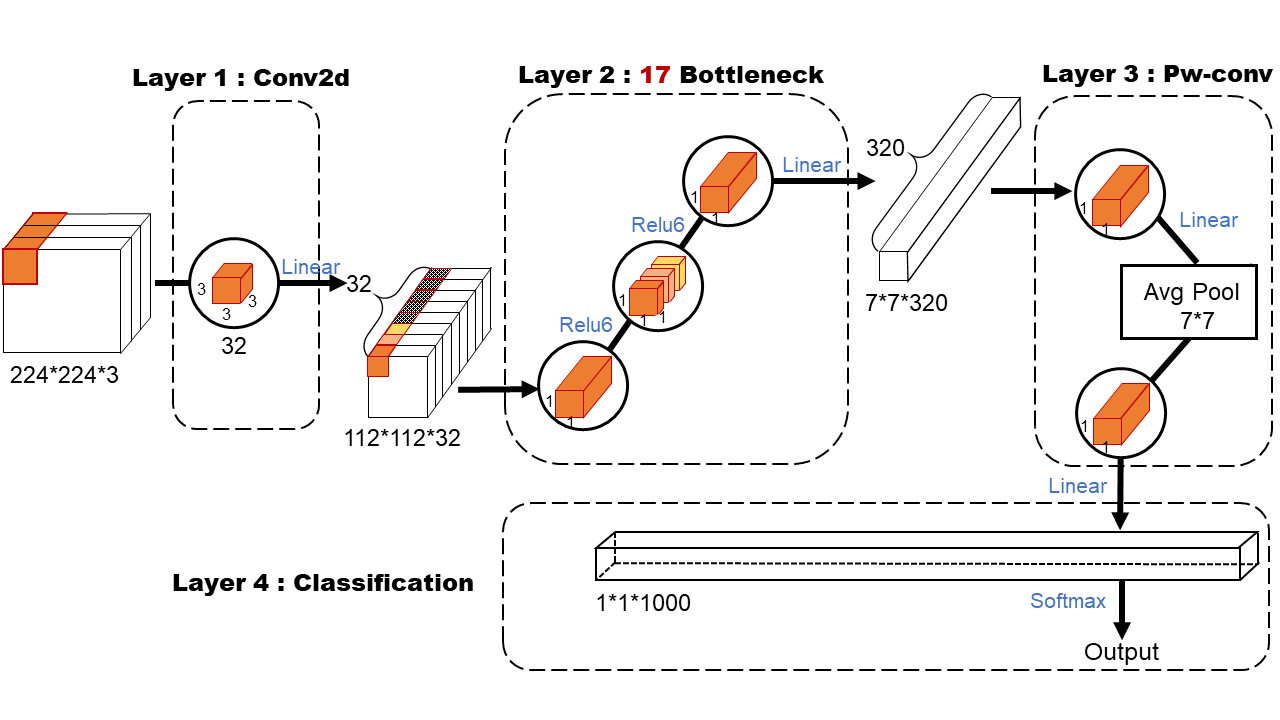}
	\caption{The lightweight network model architecture for classification learning}
	\label{fig_modelalgo}
\end{figure}

According to existing study\cite{Howard2017-fq,Sandler2018-ey}, we usd MobileNet to build an optimal OSA detection model, as shown in Fig \ref{fig_modelalgo}. The first layer is a standard convolutional layer with 32 filters of size 3x3x3, which undergo batch normalization and ReLU activation before passing to the second layer. The second layer utilizes an inverted residual structure: initially, a pointwise convolution expands the dimensionality, activated by ReLU6, followed by a depthwise convolution with ReLU6 activation, and finally a pointwise convolution reduces the dimensionality with a linear activation. This layer includes 17 inverted residual blocks, also known as "bottlenecks." The addition of ReLU6 minimizes the loss of high-dimensional information post-convolution. In the third layer, feature map expansion is achieved via pointwise convolution, followed by dimensionality reduction using an average pooling layer to prevent overfitting. A final 1x1 pointwise convolution adjusts the number of channels in the feature map. In the fourth layer, softmax activation is applied to the fully connected layer for classification predictions, where all neurons are fully connected and learning occurs through forward and backpropagation algorithms.

Note that after each convolution (either normal convolution, deep convolution or point-by-point convolution) a batch normalization and activation function operation is performed, which is omitted in Fig \ref{fig_modelalgo}.

\section{Results}
\subsection{Experiment setup}

The deep learning framework selected for this experiment is Huawei's self-developed MindSpore. During model training, the Adam optimizer is used. The batch size is set to 32, and the number of epochs is set to 20. The ECG data from 25 subjects in the UCDDB dataset is uniformly classified into two states: Sleep (S) and Non-Sleep (NS), and then input into the MobileNetV2 model for training. Subsequently, the overnight data from each subject in the Apnea-ECG and MIT-BIH datasets is fed into the model for sleep duration prediction. Finally, a rule-based screening algorithm designed for respiratory signals in the MATLAB platform is used to compute the AHI index for each subject, which is then compared with the AHI values provided in the datasets. More details can be found in Table \ref{tab_data0}.

\begin{table}[h]
	
	\centering
	\caption{Distribution of Sleep segments in three database}
	\label{tab_data0}
	\renewcommand\arraystretch{1.2}
	\begin{tabular}{@{}cccc@{}}
		{Number} & {Train set (UCDDB)}  & {MIT-BIH}  & {Apnea-ECG}\\
		\hline
		\centering
		Sleep & 13614 & 6268  &\multirow{2}{*}{7466}\\
		Non-Sleep & 5974 & 3680	\\
		
		\hline
		\multicolumn{3}{p{200pt}}{}
	\end{tabular}
	
\end{table}

\subsection{Performance evaluation}
We use statistical values (like accuracy, precision, recall, specificity and Roc) to quantify the performance of our model. The formulas for these parameters are shown in Equation \ref{eq:0}, \ref{eq:1}, \ref{eq:2}, \ref{eq:3}, \ref{eq:4}, where TP, FP, TN, and FN are true positive, false positive, true negative, and false negative, respectively.

\begin{equation}
	Accuracy = \frac{TP+TN}{TP+FP+TN+FN}
	\label{eq:0}
\end{equation}

\begin{equation}
	Precision = \frac{TP}{TP+FP}
	\label{eq:1}
\end{equation}

\begin{equation}
	Recall = \frac{TP}{TP+FN}
	\label{eq:2}
\end{equation}

\begin{equation}
	Specificity = \frac{TN}{TN+FP}
	\label{eq:3}
\end{equation}

\begin{equation}
	F1-score = 2 \times \frac{Precision\times Recall}{Precision + Recall}
	\label{eq:4}
\end{equation}

\subsection{Classification performances}
\begin{figure}[h]
	\centering
	\subfigure[Confusion Matrix of Apnea-ECG dataset(8) about SAS]
	{\label{fig_ap1-1}
		\includegraphics[width=0.4\columnwidth]{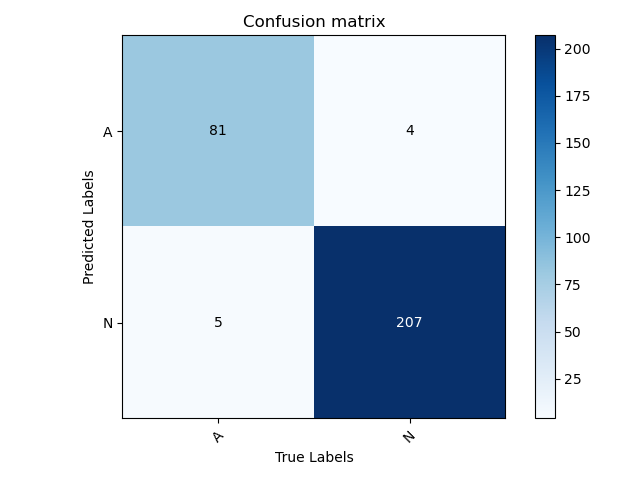}}
	\hfil
	\subfigure[ROC Curve of Apnea-ECG dataset(8) about SAS]
	{\label{fig_ap1-2}
		\includegraphics[width=0.4\columnwidth]{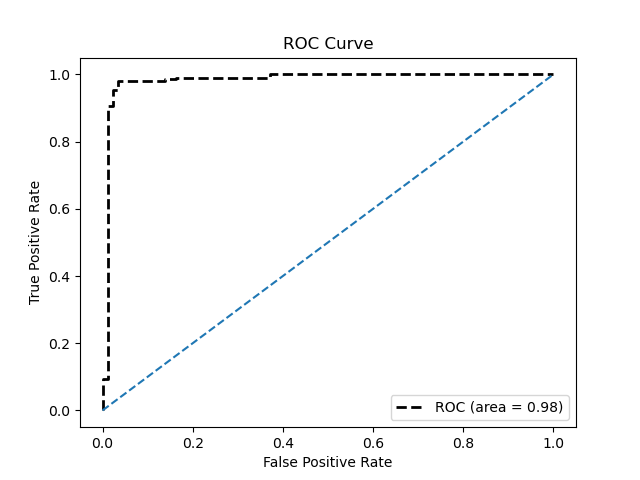}}
	\\
    \subfigure[Confusion Matrix of Apnea-ECG dataset(50) about SAS]
	{\label{fig_ap1-1}
		\includegraphics[width=0.4\columnwidth]{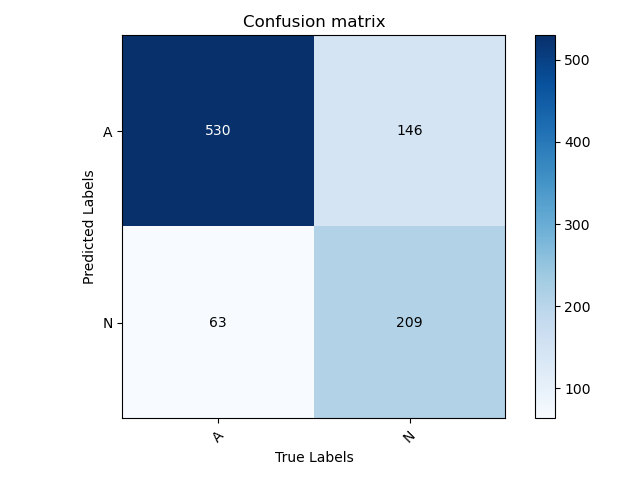}}
	\hfil
	\subfigure[ROC Curve of Apnea-ECG dataset(50) about SAS]
	{\label{fig_ap1-2}
		\includegraphics[width=0.4\columnwidth]{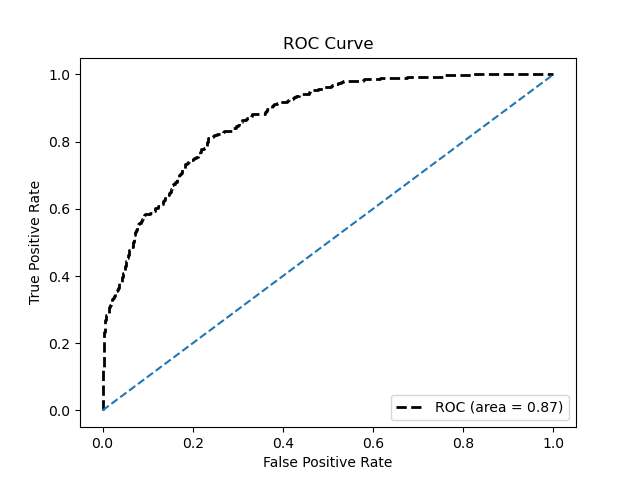}}
	\caption{Model classification results in Apnea-ECG dataset}
	\label{fig_ap1}
\end{figure}

In this study, we performed the detection of respiratory abnormal events using data from the Apnea-ECG dataset, specifically focusing on two different sample sizes: a small sample of 8 individuals and a larger sample of 50 individuals. For each frame, which corresponds to a 1-minute segment of the Electrocardiogram-derived Respiratory (EDR) signal spectrogram, the classification results of the model are shown in Fig \ref{fig_ap1}. Since the Apnea-Hypopnea Index (AHI) is considered the gold standard for diagnosing sleep apnea syndromes (SAS) and other sleep-related breathing disorders, its accurate computation is crucial. The calculation of AHI relies heavily on the precise measurement of sleep duration, as defined by Equation \ref{eq:AHI}.

\begin{equation}
	AHI = \frac{num}{Sleep Time(/h)}
	\label{eq:AHI}
\end{equation}

\begin{figure}[h]
	\centering
	\subfigure[Confusion Matrix of UCDDB about SAS]
	{\label{fig_ucddb1-1}
		\includegraphics[width=0.4\columnwidth]{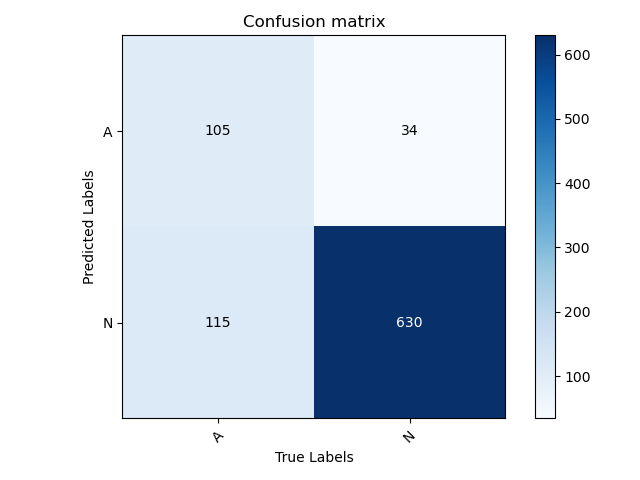}}
	\hfil
	\subfigure[ROC Curve of UCDDB about SAS]
	{\label{fig_ucddb1-2}
		\includegraphics[width=0.4\columnwidth]{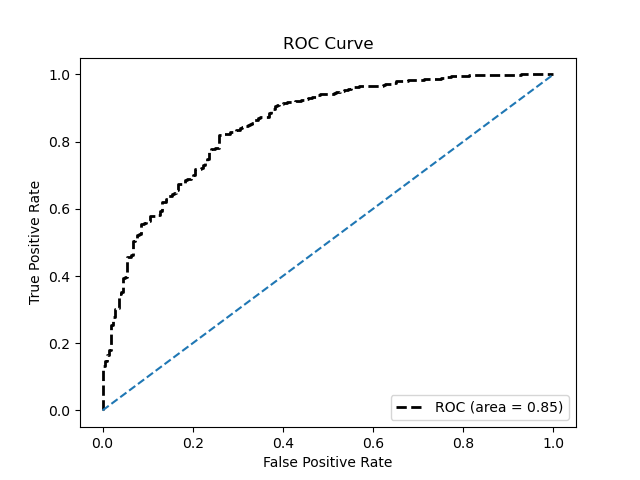}}
	\\
    \subfigure[Confusion Matrix of UCDDB about Sleep Time]
	{\label{fig_ucddb1-1}
		\includegraphics[width=0.4\columnwidth]{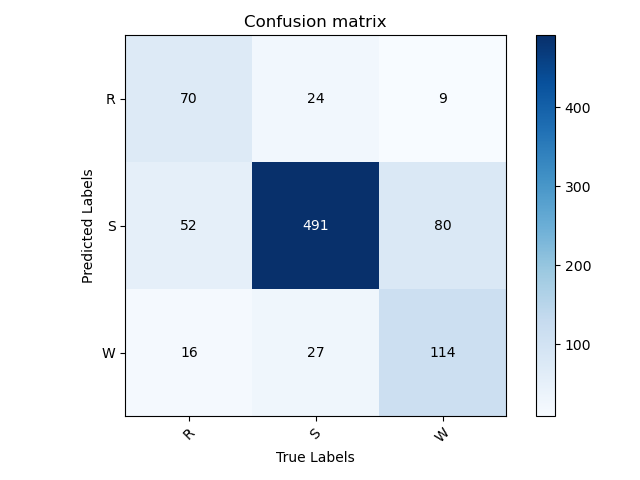}}
	\hfil
	\subfigure[ROC Curve of of UCDDB about Sleep Time]
	{\label{fig_ucddb1-2}
		\includegraphics[width=0.4\columnwidth]{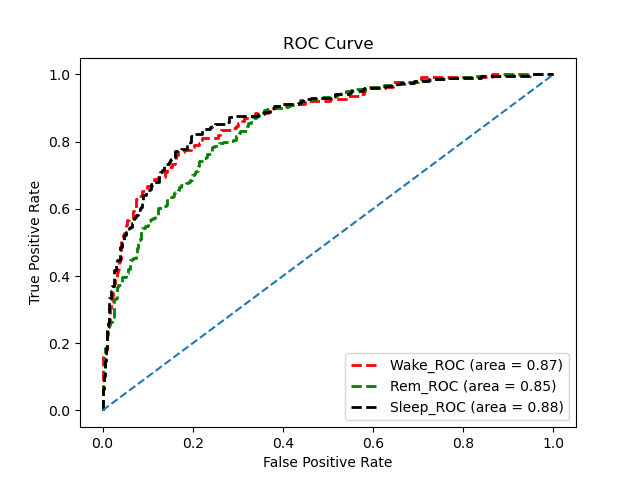}}
	\caption{Model classification results in UCDDB dataset}
	\label{fig_ucddb1}
\end{figure}

To enhance the prediction accuracy, we employed the UCDDB dataset, applying our model to perform sleep stage classification and normal/abnormal respiratory status identification during sleep. Specifically, we performed a three-class prediction to differentiate the stages of sleep and a binary classification to distinguish between normal and abnormal respiratory events. The results of these predictions are presented in Fig \ref{fig_ucddb1}. The model demonstrated strong performance in recognizing respiratory states in each frame, as well as accurately identifying sleep stages.

\begin{figure}[h]
	
	\centering
	\subfigure[Four classification prediction results]
	{\label{fig_cm1-1}
		\includegraphics[width=0.4\columnwidth]{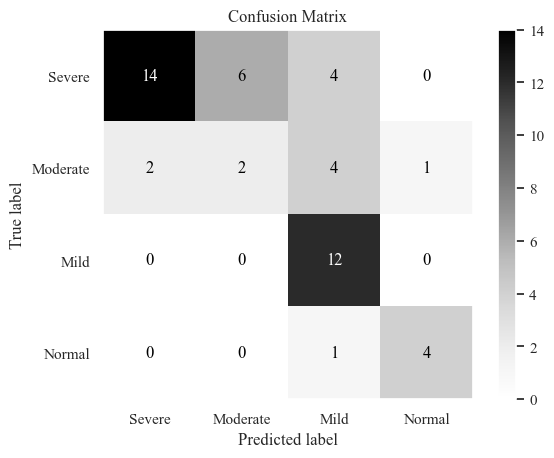}}
	\hfil
	\subfigure[Two classification prediction results]
	{\label{fig_cm1-2}
		\includegraphics[width=0.4\columnwidth]{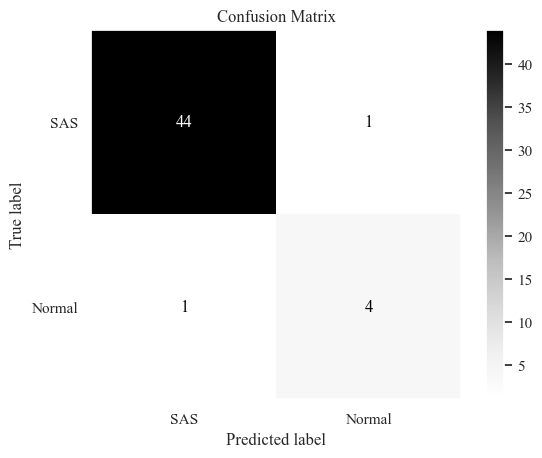}}
	
	\caption{Confusion matries of OSA degree (based on MindSpore)}
	\label{fig_cm1}
\end{figure}

\begin{figure}[h]
	
	\centering
	\subfigure[Comparison plot of AHI in Apnea-ECG dataset]
	{\includegraphics[width=0.6\columnwidth]{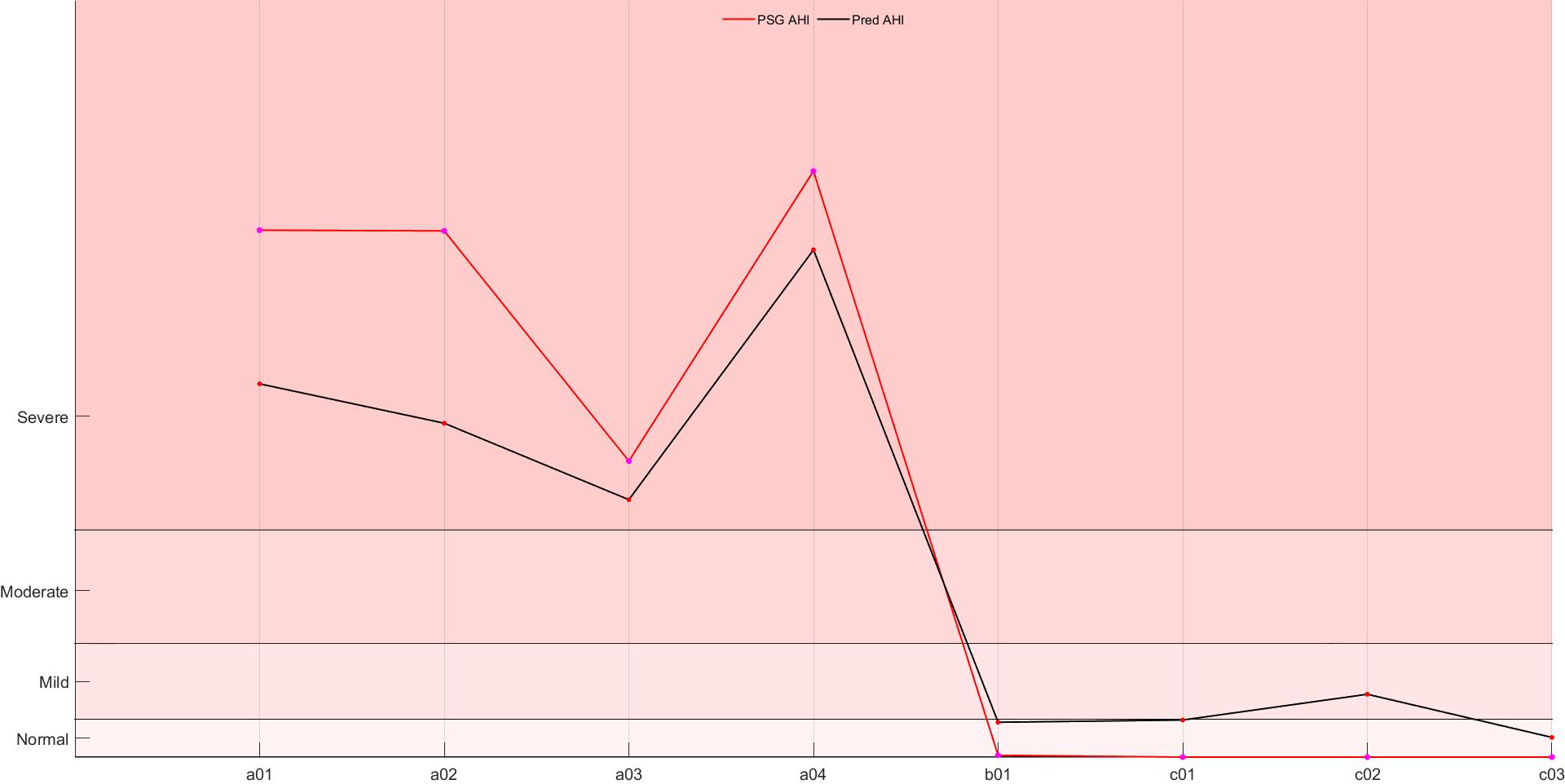}}
	\\
	\subfigure[Comparison plot of AHI in MIT-BIH dataset]
	{\includegraphics[width=0.6\columnwidth]{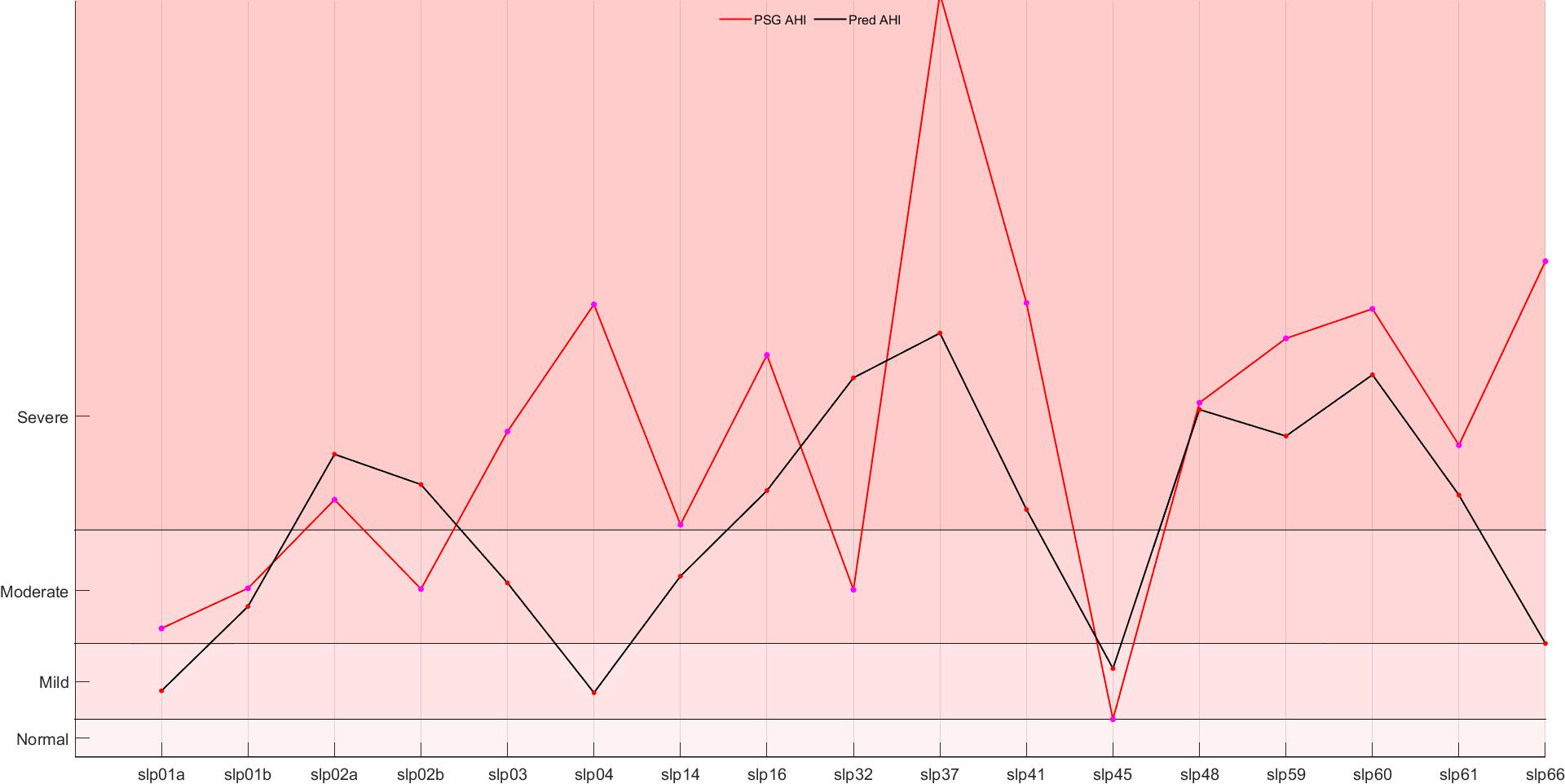}}
	\\
	\subfigure[Comparison plot of AHI in UCDDB dataset]
	{\includegraphics[width=0.6\columnwidth]{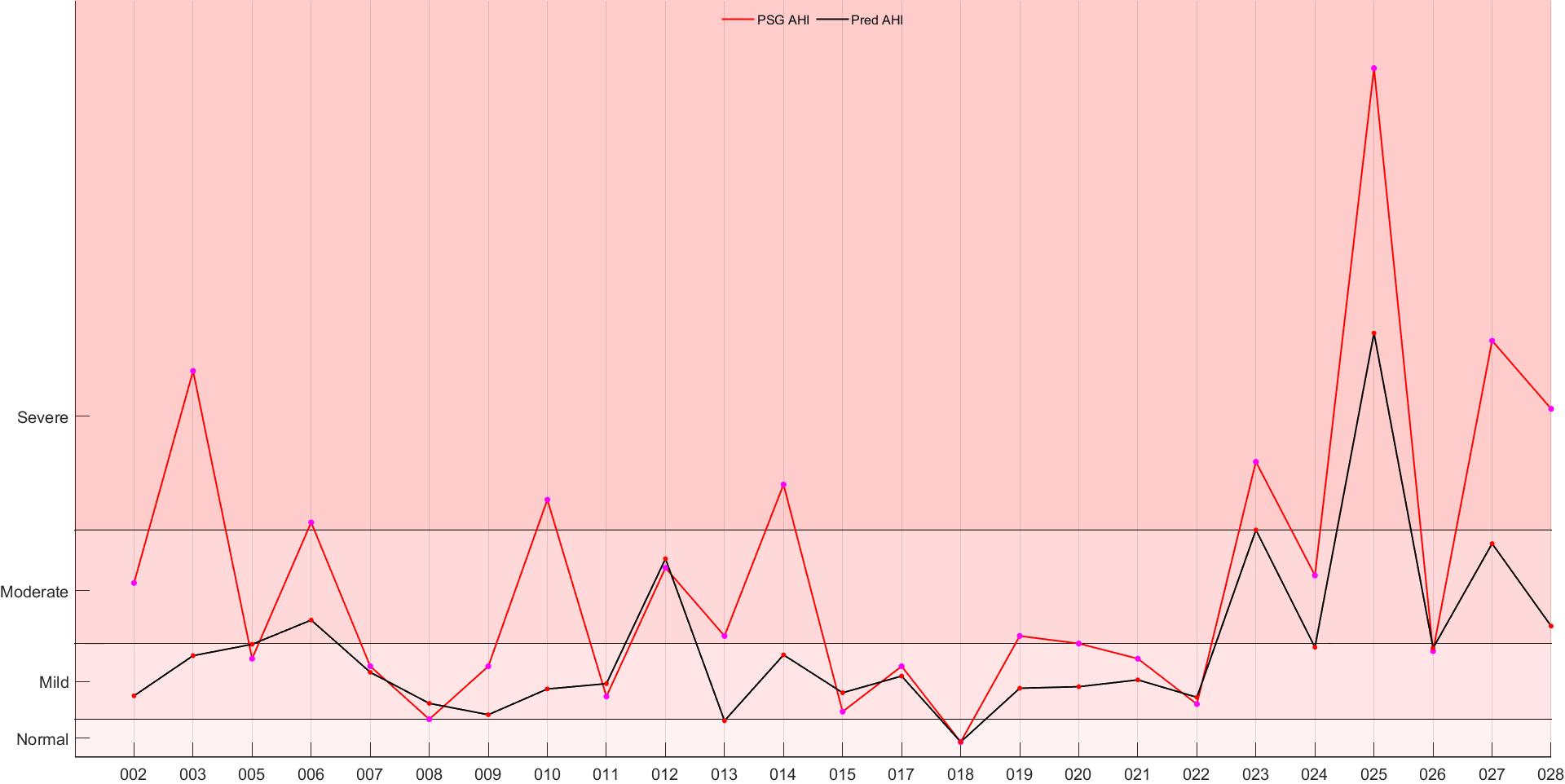}}

	\caption{Real vs. Predicted AHI for UCDDB, Apnea-ECG, and MIT-BIH Datasets}
	\label{fig_ahi_line}
\end{figure}

Given that early screening tools are typically used for interpreting full-night data for individual users, there is inherent variability between individuals in terms of sleep patterns and breathing characteristics. To address this variability, we further expanded the dataset by incorporating the MIT-BIH database, dividing the data by individual subjects rather than categorizing it solely by disease type. After training the model using data from specific subjects, the prediction results for each individual are shown in Fig \ref{fig_cm1}, \ref{fig_ahi_line}. These results indicate that our model is capable of accurately and promptly providing personalized disease risk assessments, tailored to individual users. This is particularly significant for personalized health monitoring and early diagnosis, as it reflects the model's ability to generalize across diverse user profiles.

The four-class classification task aims to categorize the AHI index into the following classes: Normal (AHI\textless 5), Mild (5 $\leq$ AHI\textless 15), Moderate ((15 $\leq$ AHI\textless 30), and Severe (AHI\textgreater 30). The confusion matrix for three datasets is presented in Fig \ref{fig_cm1-1}, where the rows represent the predicted classes, and the columns represent the true classes. The confusion matrices show how well the model differentiates between the different AHI levels.
In view of the target group of our early screening instrument, we uniformly regard those with an AHI greater than 5 as having a risk of SAS disease, and those with an AHI less than or equal to 5 as normal, as shown in Fig \ref{fig_cm1-2}.

From the confusion matrices, it can be observed that the model performs well in distinguishing between Normal and Severe classes across all datasets. However, misclassifications are more frequent in the Mild and Moderate classes, where there are more off-diagonal entries, suggesting a challenge in differentiating between these intermediate AHI levels.

The performance of the model is quantitatively assessed using the following metrics: Accuracy (ACC), Recall, Precision, F1-score, and the Macro-average of these metrics across all four classes. The results are summarized in Table \ref{tab_data1}.

\begin{table}[h]
	\centering
	\caption{The confusion matrix for three datasets about OSA classification  (based on MindSpore)}
	\label{tab_data1}
	\renewcommand\arraystretch{1.2}
	\begin{tabular}{@{}c|ccccc@{}}
		\hline
		{} & {Precision}  & {Recall}  & {Specificity}  & {F1-score} &{Acc}\\
		\hline
		\centering
		SAS & 0.978 & 0.978 & 0.800 & 0.978 &\multirow{2}{*}{0.960}\\
		Non-SAS & 0.800 & 0.800 & 0.978 & 0.800 	\\
		\hline
	\end{tabular}
\end{table}


To enhance the robustness of our findings, we performed a comparative analysis with recent studies that utilized similar lightweight neural network models for obstructive sleep apnea (OSA) detection. This comparison highlights the advantages and limitations of our approach.

Our model demonstrated high classification accuracy across all datasets, achieving a mean accuracy of 0.978 for OSA detection. Precision, recall, and F1-scores were also consistently high. Specifically, for the Apnea-ECG dataset, the ROC-AUC was 0.98, underscoring the robustness of our model. This performance was compared to other state-of-the-art methods as shown in Table~\ref{tab:comparison}.

\begin{table}[h]
    \centering
    \caption{Performance comparison with recent studies on lightweight neural networks for OSA detection.}
    \label{tab:comparison}
    \begin{tabular}{|c|c|c|c|c|}
        \hline
        \textbf{Study} & \textbf{Model} & \textbf{Dataset} & \textbf{Accuracy} & \textbf{ROC-AUC} \\
        \hline
          &  & Apnea-ECG(8) & 0.970 & 0.98 \\
          &  & Apnea-ECG(50) & 0.780 & 0.87 \\
        Our Study & MobileNetV2 & UCDDB(24) & 0.831 & 0.85 \\
          &  & Apnea-ECG UCDDB MIT-BIH (50) & 0.960 & -- \\
        \hline
        Novak et al. (2008)~\cite{novak2008apnea} & LSTM-Based & Apnea-ECG & 0.897 & -- \\
        \hline
        Pathinarupothi et al. (2017)~\cite{pathinarupothi2017single} & LSTM ($ SpO_{2} $ + IHR) & OSA Dataset & 0.921 & 0.99 \\
        \hline
        Li et al. (2018)~\cite{li2018detection} & ANN-HMM & Single-lead ECG & 0.889 & 0.869 \\
        \hline
        Thompson et al. (2020)~\cite{thompson2020detection} & 1D-CNN & Apnea-ECG & 0.938 & 0.9945 \\
        \hline
        Chen et al. (2022)~\cite{chen2022spatio} & CNN-BiGRU & UCDDB & 0.923 & 0.890 \\
        \hline
    \end{tabular}
\end{table}

Despite these promising results, we observed that the model's runtime remains a challenge, particularly when deployed on large-scale datasets or resource-constrained environments. To address this, quantization techniques from \textit{LLMEasyQuant}~\cite{liu2024llmeasyquant} could be adapted to optimize the computational efficiency of MobileNetV2, while dynamic graph sampling strategies from \textit{GraphSnapShot}~\cite{liu2024graphsnapshot} might aid in processing patient data that includes temporal and spatial dependencies~\cite{liu2024distance}. These advancements provide a foundation for further exploration into efficient and scalable approaches, enhancing the applicability of lightweight models in real-world healthcare scenarios.


In addition to the confusion matrices and classification metrics, we further evaluate the model's performance by comparing the real AHI values with the predicted (calculated) AHI values for each dataset. In Fig \ref{fig_ahi_line}, line plots are presented for each dataset to visualize the correspondence between the true AHI values and the model's predictions over the entire set of subjects. These plots provide an insightful way to assess the overall fit and the model's ability to approximate the true AHI levels across different stages of sleep apnea severity.

The real vs. predicted AHI line plots serve as a visual confirmation of the model's predictive capabilities. The strong alignment between the true and predicted values across all datasets indicates that the model is effective in approximating the AHI values with high accuracy. Although some deviations are observed, particularly in the Mild and Moderate AHI categories, these can be attributed to the inherent challenges in distinguishing between these similar levels of severity. Overall, the line plot analysis reinforces the favorable results obtained from the confusion matrix and evaluation metrics, suggesting that the model is highly effective for the task of sleep apnea classification based on AHI prediction.

\begin{figure}[h]
	\centering
	\includegraphics[width=0.8\columnwidth]{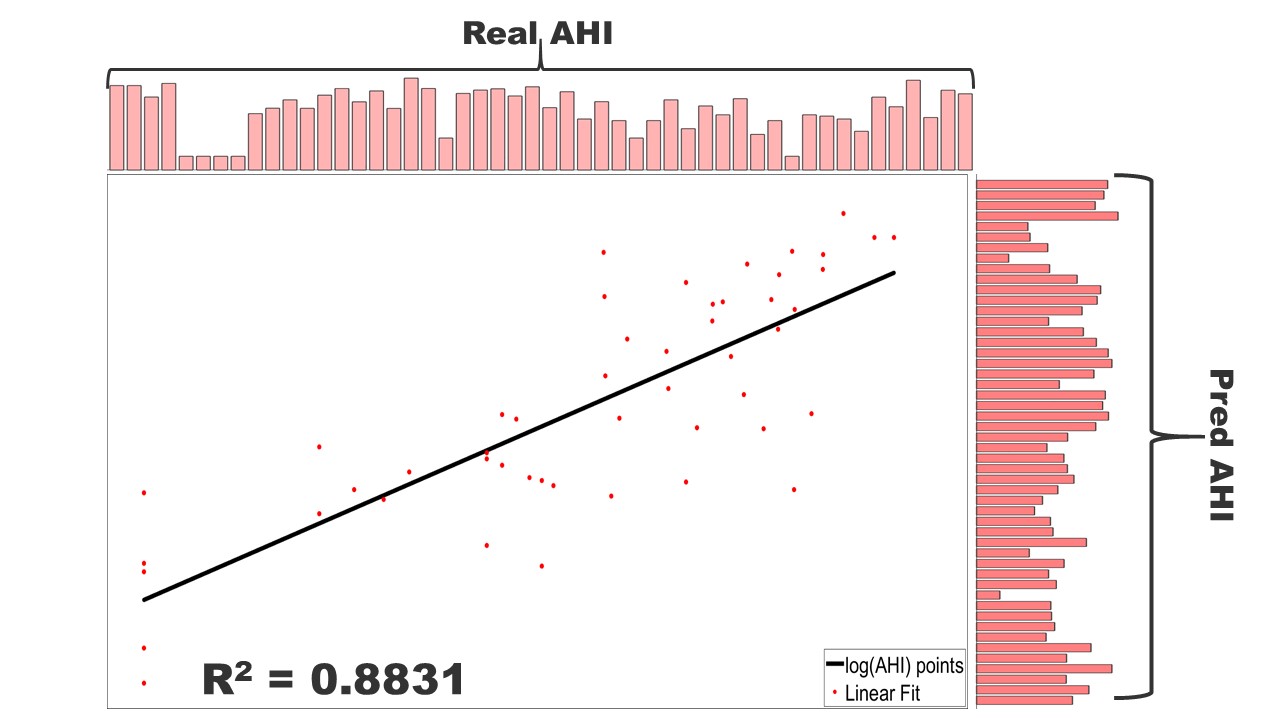}
	\caption{Comparsion}
	\label{fig_result50}
\end{figure}

The integration of predicted and actual AHI (Apnea-Hypopnea Index) data from three publicly available databases enabled us to construct comprehensive AHI prediction and ground truth curves for a cohort of 50 individuals, as Fig \ref{fig_result50}. Statistical analysis revealed a strong correlation between the predicted and actual curves, with a Pearson correlation coefficient of 0.8831. This high degree of concordance underscores the robustness of the proposed model in accurately predicting AHI values across diverse datasets and individual cases, demonstrating its potential applicability in clinical and wearable health-monitoring settings.

\section{Conclusion}
The proposed machine learning model demonstrates promising results for the classification of AHI indices in the context of sleep apnea screening. The confusion matrices reveal that the model is generally effective at identifying Normal and Severe classes, but it encounters more difficulty in distinguishing between the Mild and Moderate categories, likely due to the similarities in these intermediate states. The overall accuracy, recall, precision, and F1-score metrics further validate the model's robustness in the task of sleep apnea classification.

The results of this study have important implications, especially in improving the efficiency of early screening for sleep apnea syndrome (OSA). By combining blood oxygenation and ECG signals and utilizing a lightweight neural network model, this approach is not only able to rival traditional complex models (e.g., PSG) in terms of accuracy, but also has the potential to be practically applied to smart wearable devices. This could help to significantly reduce screening costs and improve the current low diagnosis rate of OSA due to expensive equipment and insufficient medical resources. Meanwhile, the method can also provide a new technical route for other disease detection and sleep health monitoring based on ECG and blood oxygen signals.


Our MobileNetV2 model achieved superior accuracy and ROC-AUC metrics. This improvement is attributable to the integration of ECG and respiratory signal features, along with advanced preprocessing techniques to mitigate noise artifacts. While Howard et al.\cite{2017MobileNets} leveraged a similar architecture, their results on the MIT-BIH dataset fell short due to less comprehensive signal integration. However, certain limitations should be acknowledged. Misclassification rates were higher in the mild and moderate OSA categories, suggesting room for improvement in feature extraction and training methodologies. Future work will explore hybrid models combining convolutional and recurrent architectures to address these challenges.


\backmatter

\bmhead{Supplementary information}
None.

\bmhead{Acknowledgements}
The authors acknowledge all the participants and survey staffs for their participation. The work in this paper was carried out as a Shenzhen Science and Technology Major Project 'Heavy 2023 24N122 Key Technology Research and Development of Intelligent Sleep Monitoring and Analysis System'. In addition, the authors of the paper give extra thanks to Huawei's open source AI framework MindSpore. Our code are available at \href{https://github.com/mindspore-lab/models/tree/master/research/arxiv_papers/Easy-MobileNetV2}
{https://github.com/mindspore-lab/models/tree/master/research/arxiv\_papers/Easy-MobileNetV2}.

\section*{Declarations}

\begin{itemize}
\item Funding
This study was funded by the Shenzhen Science and Technology Major Project (KJZD20230923114306013); 

\item Conflict of interest/Competing interests
None.

\item Ethics approval and consent to participate
The study was conducted according to the World Medical Association 
Declaration of Helsinki in 1975, as revised in 1983, and was approved by the 
Ethic Committee of Shanghai Jiao Tong University Affiliated Sixth People's 
Hospital (Trial registration number: ChiCTR1900025714). All subjects provided 
their informed written consent

\item Consent for publication
Not applicable.

\item Data availability 
No datasets were generated or analysed during the current study.

\item Author contribution
The authors take responsibility and vouch for the accuracy and completeness 
of the data and analyses. Prof. ZX, YJL had full access to all of the data in 
the study and took responsibility for the integrity of the data and the accuracy 
of the data analysis. Study design: ZX; Data collection:PH; Statistical analysis: PH, YYX; Manuscript draft: PH, YYX, ZX. The authors have seen and approved the manuscript

\end{itemize}

\bibliography{sleep-article_ph.bib}

\end{document}